\documentclass{article}

\usepackage[final]{neurips_2023}

\usepackage[utf8]{inputenc} 
\usepackage[T1]{fontenc}    
\usepackage{hyperref}       
\usepackage{url}            
\usepackage{booktabs}       
\usepackage{amsfonts}       
\usepackage{nicefrac}       
\usepackage{microtype}      
\usepackage{xcolor}         
\usepackage{amsmath}
\usepackage{graphicx}
\newtheorem{definition}{Definition}
\title{Asymmetric Norms to Approximate the Minimum Action Distance}

\author{%
  Lorenzo Steccanella, Anders Jonsson\\
  Dept. Information and Communication Technologies, \\
  Universitat Pompeu Fabra, Barcelona, Spain \\
  \texttt{\{lorenzo.steccanella, anders.jonsson\}@upf.edu} \\
}

\begin{document}

\maketitle

\begin{abstract}
    This paper presents a state representation for reward-free Markov decision processes. The idea is to learn, in a self-supervised manner, an embedding space where distances between pairs of embedded states correspond to the minimum number of actions needed to transition between them. Unlike previous methods, our approach incorporates an asymmetric norm parametrization, enabling accurate approximations of minimum action distances in environments with inherent asymmetry. We show how this representation can be leveraged to learn goal-conditioned policies, providing a notion of similarity between states and goals and a useful heuristic distance to guide planning.

    To validate our approach, we conduct empirical experiments on both symmetric and asymmetric environments. Our results show that our asymmetric norm parametrization performs comparably to symmetric norms in symmetric environments and surpasses symmetric norms in asymmetric environments.
\end{abstract}

\section{Minimum Action Distance} \label{sec:Minimum Action Distance}

In this section, we describe the notion of Minimum Action Distance, and we derive useful ways of computing this measure on finite MDPs and continuous or large MDPs.

We start by introducing some notation. 


\begin{definition}
    The Minimum Action Distance (MAD) $d_{MAD}: \mathcal{S}^2 \rightarrow \mathbb{R}^+$ is defined as the minimum number of decision steps to transition between any pair of states $(s_i, s_j) \in \mathcal{S}^2$.
\end{definition}

We can observe that the MAD is an asymmetric distance function \citep{mennucci2013asymmetric} and must satisfy the following properties:

\begin{itemize}
    \item $d_{MAD} \geq 0$ and $\forall s \in \mathcal{S}$, $d_{MAD}(s, s) = 0$.
    \item $d_{MAD}(s_i, s_j) = d_{MAD}(s_j, s_i) = 0$ implies $s_i = s_j$.
    \item $d_{MAD}(s_i, s_j) \leq d_{MAD}(s_i, s_k) + d_{MAD}(s_k, s_j)$ $\forall (s_i, s_j, s_k) \in \mathcal{S}^3$.
\end{itemize}

\subsection{Learning Minimum Action Distance from Adjacency Matrix}

In discrete and finite MDPs we can compute the state-transition graph $G=(V, E)$ of an MDP. In this section, we will revise how to learn the minimum action distance from the graph adjacency matrix.

A state-transition graph $G=(V, E)$ of an MDP $\mathcal{M} = \langle \mathcal{S},\mathcal{A},\mathcal{P},r \rangle$ is a graph with nodes representing the states in the MDP and the edges representing state adjacency in the MDP. More precisely, $V=\mathcal{S}$, $e\left(s_i, s_j\right) \in E \,$ iff $\, \exists a \: \mathcal{P}\left(s_i, a, s_j\right)>0$. An adjacency matrix represents a graph with a square matrix of size $|\mathcal{S}| \times|\mathcal{S}|$ with $(i, j)$-value being 1 if $e\left(s_i, s_j\right) \in E$ and 0 otherwise.

\begin{equation}
    A_{i j}^{G}=\left\{\begin{array}{ll}
        0 & s_i=s_j \quad or \quad e(s_i, s_j) \notin E \\
        1 & e(s_i, s_j) \in E
    \end{array} \quad i, j=1, \ldots, |\mathcal{S}| .\right.
\end{equation}

Having access to the adjacency matrix $A^G$ we can simply compute the minimum action distance by using the Floyd-Warshall algorithm \citep{floyd, roy, warshall}.

The Floyd-Warshall algorithm compares all possible paths through the graph between each pair of vertices. It is able to do this with $\Theta\left(|V|^3\right)$ comparisons in a graph, even though there may be up to $\Omega\left(|V|^2\right)$ edges in the graph, and every combination of edges is tested. It does so by incrementally improving an estimate on the shortest path between two vertices until the estimate is optimal.

This dynamic programming procedure relies on having access to the edge weights, which in the case of MAD reduces to having access to the adjacency matrix $A^G$ of $a_{i,j} = 1$ when $s_i, s_j$ are connected by an edge.

Thanks to this we can define the shortest path on the $A^G$ by just first computing the base cases:

\begin{equation}
    d(s_i, s_j)=\left\{\begin{array}{lll}
        0      & s_i=s_j                                  \\
        1      & a_{i,j} = 1                              \\
        \infty & a_{i,j} = 0 \quad and \quad s_i \neq s_j
    \end{array} \quad i, j=1, \ldots, |\mathcal{S}|,\right.
\end{equation}
and subsequently computing the recursive case leveraging the triangle inequality property:

\begin{equation}
    d(s_i, s_j) = \min(d(s_i, s_j), d(s_i, s_k) + d(s_k, s_j)) \quad \forall (s_i, s_j, s_k) \in \mathcal{S}^3.
\end{equation}

Note that in case we do not have access to the adjacency matrix $A^G$ this can be retrieved by interacting with the environment by visiting all the one-step transitions $(s, s')$.

\subsection{Symmetric embeddings}

The Minimum Action Distance between states is a priori unknown and is not directly observable in continuous and/or noisy state spaces where we cannot simply enumerate the states and construct the adjacency matrix of the MDP. Instead, we will approximate it using the distances between states observed on trajectories. We introduce the notion of Trajectory Distance (TD) as follows:

\begin{definition}
    (Trajectory Distance) Given any trajectory $\tau=\{s_0, ..., s_n\} \sim \mathcal{M}$ collected in an MDP $\mathcal{M}$ and given any pair of states along the trajectory $(s_i, s_j) \in \tau$ such that $0 \leq i \leq j \leq n$, we define $d_{TD}(s_i, s_j \mid \tau)$ as
    \begin{equation}
        d_{TD}(s_i, s_j \mid \tau) = (j - i),
    \end{equation}
    i.e.~the number of decision steps required to reach $s_j$ from $s_i$ on trajectory $\tau$.
\end{definition}

We can observe that given any state trajectory $\tau=\{s_0, ..., s_n\}$, choosing any pair of states $(s_i, s_j) \in \tau$ with $0 \leq i \leq j \leq n$, their distance along the trajectory represents an upper bound of the MAD.

\begin{equation} \label{eqn:inequalityMAD}
    d_{MAD}(s_i, s_j) \leq d_{TD}(s_i, s_j \mid \tau).
\end{equation}

Given a dataset of trajectories $\mathcal{D}$ collected by any unknown behavior policy, we can retrieve the MAD $d_{MAD}$ by solving the following constrained optimization problem:

\begin{equation}
    \begin{aligned}
        \begin{split}
            \min_{\theta} \quad & \sum_{\tau\in \mathcal{D}}\sum_{(s_i,s_j)\in \tau} ( d_{\theta}(s_i, s_j) - d_{TD}(s_i, s_j \mid \tau))^2,\\
            \textrm{s.t.} \quad & d_{\theta}(s_i, s_j) \leq d_{TD}(s_i, s_j \mid \tau) \quad \forall (s_i, s_j) \in \mathcal{S}_{\mathcal{D}}^2
        \end{split}
    \end{aligned}
    \label{eqn:madconstrainedoptv1}
\end{equation}

As a first step we can leverage the triangle inequality to simplify the constrain in \ref{eqn:madconstrainedoptv1} and reduce the dependency on the quality of the trajectories in the dataset $\mathcal{D}$.

\begin{equation}
    \begin{aligned}
        \begin{split}
            \min_{\theta} \quad & \sum_{\tau\in \mathcal{D}}\sum_{(s_i,s_j)\in \tau} ( d_{\theta}(s_i, s_j) - d_{TD}(s_i, s_j \mid \tau))^2,\\
            \textrm{s.t.} \quad & d_{\theta}(s, s') \leq d_{TD}(s, s' \mid \tau) \quad \forall \tau \in \mathcal{D}, \forall (s,s')\in \tau, \\ &
            d_{\theta}(s_i, s_j) \leq d_{\theta}(s_i, s_k) + d_{\theta}(s_k, s_j), \quad \forall (s_i, s_j, s_k) \in \mathcal{S}_{\mathcal{D}}^3
        \end{split}
    \end{aligned}
    \label{eqn:madconstrainedoptv2}
\end{equation}
where $(s,s')\in \tau$ refers to a one-step transition (i.e. $d_{TD}(s, s' | \tau) = 1$) in the trajectory $\tau \in \mathcal{D}$ while $(s_i, s_j, s_k) \in \mathcal{S}_{\mathcal{D}}^3$ indicates all the combinations of $3$ states contained in the trajectory dataset $\mathcal{S}_{\mathcal{D}}$.

Note that the first constraint in \ref{eqn:madconstrainedoptv2} imposes an upper bound on one-step transitions, i.e. it says that two states $(s, s')$ at distance one along a trajectory $d_{TD}(s, s' | \tau) = 1$ are either the same state $s = s'$ or they must satisfy $d_{MAD} = 1$. This allows us to approximate the MAD without having to identify whether two states along a trajectory are the same state or not.

Note that the second constraint in \ref{eqn:madconstrainedoptv2} corresponds to the triangle inequality, which thus is a property that holds for the MAD.
Moreover, this second constraint implies that we have to calculate it for all the combinations of $3$ states contained in $\mathcal{S}_{\mathcal{D}}$ which can become intractable for large state spaces.

To address this issue we proposed an alternative formulation based on embedding the MAD in a parametric embedding space $\phi_\theta: \mathcal{S} \rightarrow \mathbb{R}^{dim}$ where a chosen distance metric that respects the triangle inequality (e.g. any norm $||\cdot||_p$) can be used to enforce the triangle inequality constraint.

The goal is to learn a parametric state embedding $\phi_\theta: \mathcal{S} \rightarrow \mathbb{R}^{dim}$ such that the distance $d$ between any pair of embedded states approximates the Minimum Action Distance.

\cite{steccanella2022state} used this formulation to favour symmetric embeddings, where they use norms as distance functions, e.g.~the L1 norm $d(z,y)=||z-y||_1$.
If we use symmetric embeddings we will have that for any pair of states $(s_i, s_j) \in \mathcal{S}$,

\begin{equation}
    d(\phi_\theta(s_i), \phi_\theta(s_j)) \approx \min(d_{MAD}(s_i, s_j), d_{MAD}(s_j, s_i)).
\end{equation}

The problem of learning this embedding can then be formulated as a constrained optimization problem:

\begin{equation}
    \begin{aligned}
        \min_{\theta} \quad & \sum_{\tau\in\mathcal{D}}\sum_{(s_i,s_j)\in \tau} (\left\|\phi_\theta(s_i)-\phi_\theta(s_j)\right\|_p - d_{TD}(s_i, s_j \mid \tau))^2,    \\
        \textrm{s.t.} \quad & \left\|\phi_\theta(s)-\phi_\theta(s')\right\|_p \leq d_{TD}(s, s' \mid \tau) \;\;\; \forall \tau \in \mathcal{D}, \forall (s,s')\in \tau. \\
    \end{aligned}
    \label{eqn:pf}
\end{equation}

Intuitively, the objective is to make the embedded distance between pairs of states as close as possible to the observed trajectory distance while respecting the upper bound constraints. Without constraints, the objective is minimized when the embedding matches the expected Trajectory Distance $\mathbb{E}\left[d_{TD}\right]$ between all pairs of states observed on trajectories in the dataset $\mathcal{D}$. In contrast, constraining the solution to match the minimum TD with the upper-bound constraints $\left\|\phi_\theta(s)-\phi_\theta(s')\right\|_p \leq d_{TD}(s, s' \mid \tau)$ allows us to approximate the MAD. The precision of this approximation depends on the quality of the given trajectories.

To make the constrained optimization problem tractable, we can relax the hard constraints in \eqref{eqn:pf} and convert them into a penalty term to retrieve a simple unconstrained formulation. Moreover, we rely on sampling $(s_i, s_j, d_{TD}(s_i, s_j \mid \tau))$ and $(s, s', d_{TD}(s, s' \mid \tau))$ from the dataset of trajectories $\mathcal{D}$ making this formulation amenable for gradient descent and to fit within the optimization scheme of neural networks.

\begin{equation} \label{eqn:relaxed_pf}
    \begin{aligned}
        \resizebox{0.9\hsize}{!}{$\mathcal{L} = \mathbb{E}_{(s_i, s_j, d_{TD}(s_i, s_j \mid \tau)) \sim \mathcal{D}} \left[ (\left\|\phi_\theta(s_i)-\phi_\theta(s_j)\right\|_p - d_{TD}(s_i, s_j \mid \tau))^2\right] + C,$}
    \end{aligned}
\end{equation}
where $C$ is our penalty term defined as
\begin{equation} \label{eqn:penalty_term}
    \begin{aligned}
        \resizebox{0.9\hsize}{!}{$C = \mathbb{E}_{(s, s', d_{TD}(s, s' \mid \tau)) \sim \mathcal{D}} \left[ \max \left(0, \left\|\phi_\theta(s)-\phi_\theta(s')\right\|_p - d_{TD}(s, s' \mid \tau) \right)^2\right]$}.
    \end{aligned}
\end{equation}

The penalty term $C$ introduces a quadratic penalization of the objective for violating the upper-bound constraints $\left\|\phi_\theta(s)-\phi_\theta(s')\right\|_p <= d_{TD}(s, s' \mid \tau)$.

\subsection{Asymmetric embeddings}

In the previous section, we have seen how it is possible to define the MAD embedding problem with the use of norms $|| \cdot ||_p$. While the formulation is useful to understand how it is possible to remove the triangle inequality constraint in \ref{eqn:madconstrainedoptv2} the Minimum Action Distance is naturally asymmetric and we would like embedding that preserves this asymmetry.

A norm is a function $\|\cdot\|: \mathcal{X} \rightarrow \mathbb{R}$ satisfying, $\forall x, y \in \mathcal{X}, \alpha \in \mathbb{R}^{+}$:
\begin{itemize}
    \item $\mathbf{N 1}$ (Pos. def.). $\|x\|>0$, unless $x=0$.
    \item $\mathbf{N 2}$ (Pos. homo.). $\alpha\|x\|=\|\alpha x\|$, for $\alpha \geq 0$.
    \item $\mathbf{N 3}$ (Subadditivity). $\|x+y\| \leq\|x\|+\|y\|$.
    \item $\mathbf{N 4}$ (Symmetry). $\|x\|=\|-x\|$.
\end{itemize}

An asymmetric semi-norm satisfies $\mathbf{N 2}$, $\mathbf{N 3}$ but not necessarily $\mathbf{N 1}$, $\mathbf{N 4}$.

A convex function $f: \mathcal{X} \rightarrow \mathbb{R}$ is a function satisfying $\mathbf{C 1}: \forall x, y \in \mathcal{X}, \alpha \in[0,1]: f(\alpha x+(1-\alpha) y) \leq$ $\alpha f(x)+(1-\alpha) f(y)$. The commonly used ReLU activation, $\operatorname{relu}(x)=\max (0, x)$, is convex.

Is easy to observe that any $\mathbf{N2}$ and any $\mathbf{N3}$ function is convex and thus that any asymmetric semi-norm is convex.

Motivated by this relationship between convex functions and norms,\cite{pitis2020inductive} introduced Wide Norms, a parametric distance that models symmetric and asymmetric norms.

A Wide Norm is any combination of symmetric/asymmetric semi-norms. They are based on the Mahalanobis norm of $x \in \mathbb{R}^{dim}$, parametrized by $W \in \mathbb{R}^{m \times n}$, defined as $||x||_W = ||Wx||_2$.

Asymmetric Wide Norms are defined as:

\begin{equation*}
    ||x|_{WN}= || W relu(x :: -x) ||_2 \text{   where   } W_i \in \mathbb{R}^{m_i \times n} \text { with } m_i \leq n \text {. }
\end{equation*}

We can then use the parametrized Wide Norm distance to constrain the triangular inequality on the embedding space:

\begin{equation} \label{eqn:relaxed_pf_asym}
    \begin{aligned}
        \resizebox{0.9\hsize}{!}{$\mathcal{L} = \mathbb{E}_{(s_i, s_j, d_{TD}(s_i, s_j \mid \tau)) \sim \mathcal{D}} \left[ (||\phi_\theta(s_i)-\phi_\theta(s_j)|_{WN} - d_{TD}(s_i, s_j \mid \tau))^2\right] + C,$}
    \end{aligned}
\end{equation}
where $C$ is our penalty term defined as
\begin{equation} \label{eqn:penalty_term_asym}
    \begin{aligned}
        \resizebox{0.9\hsize}{!}{$C = \mathbb{E}_{(s, s', d_{TD}(s, s' \mid \tau)) \sim \mathcal{D}} \left[ \max \left(0, ||\phi_\theta(s)-\phi_\theta(s')|_{WN} - d_{TD}(s, s' \mid \tau) \right)^2\right]$}.
    \end{aligned}
\end{equation}


\section{Learning the Transition Model}

In the previous section, we showed how to learn a state representation that encodes a distance metric between states. This distance allows us to measure how far we are from the goal state, i.e.~$d(\phi_\theta(s_t), \phi_\theta(s_{goal}))$. However, on its own, the distance metric does not directly give us a policy for reaching the desired goal state.

In this section we propose a method to learn a transition model of actions, that combined with our state representation allows us to plan directly in the embedded space and derive policies to reach any given goal state. Given a dataset of trajectories $\mathcal{T}$ and a state embedding $\phi_\theta(s)$, we seek a parametric transition model $\rho_\zeta(\phi_\theta(s), a)$ such that for any triple $(s, a, s') \in \mathcal{T}$, $\rho_\zeta(\phi_\theta(s), a) \approx \phi_\theta(s')$. 

We propose to learn this model simply by minimizing the squared error as:

\begin{equation}
\begin{aligned}
\min_{\zeta} \quad & \sum_t^{\mathcal{D}}\sum_{s, a, s'}^{t} \left[ (\rho_\zeta(\phi_\theta(s), a) - \phi_\theta(s'))^2\right].\\
\end{aligned}
\label{eqn:transition_model}
\end{equation}

Once we learned the model, given a goal $g$ we can select actions that minimize the $d_{MAD}$ distance to the goal in the latent space $\phi_\theta$.

\begin{equation*}
    \arg\min_{a \in A} ||\rho_\zeta(\phi_\theta(s), a) - \phi_\theta(g)|_{WN}
\end{equation*}

\section{Empirical Evaluation}

In this section we report a comparison of symmetric norms and WideNorms performance on symmetric and asymmetric environments. 

We refer to $L1$ as the agent that learns the $d_{MAD}$ using an L1 norm and to $WideNorm$ agent the agent that uses WideNorms to approximate the $d_{MAD}$.

\begin{figure}
    \centering
    \includegraphics[width=\textwidth]{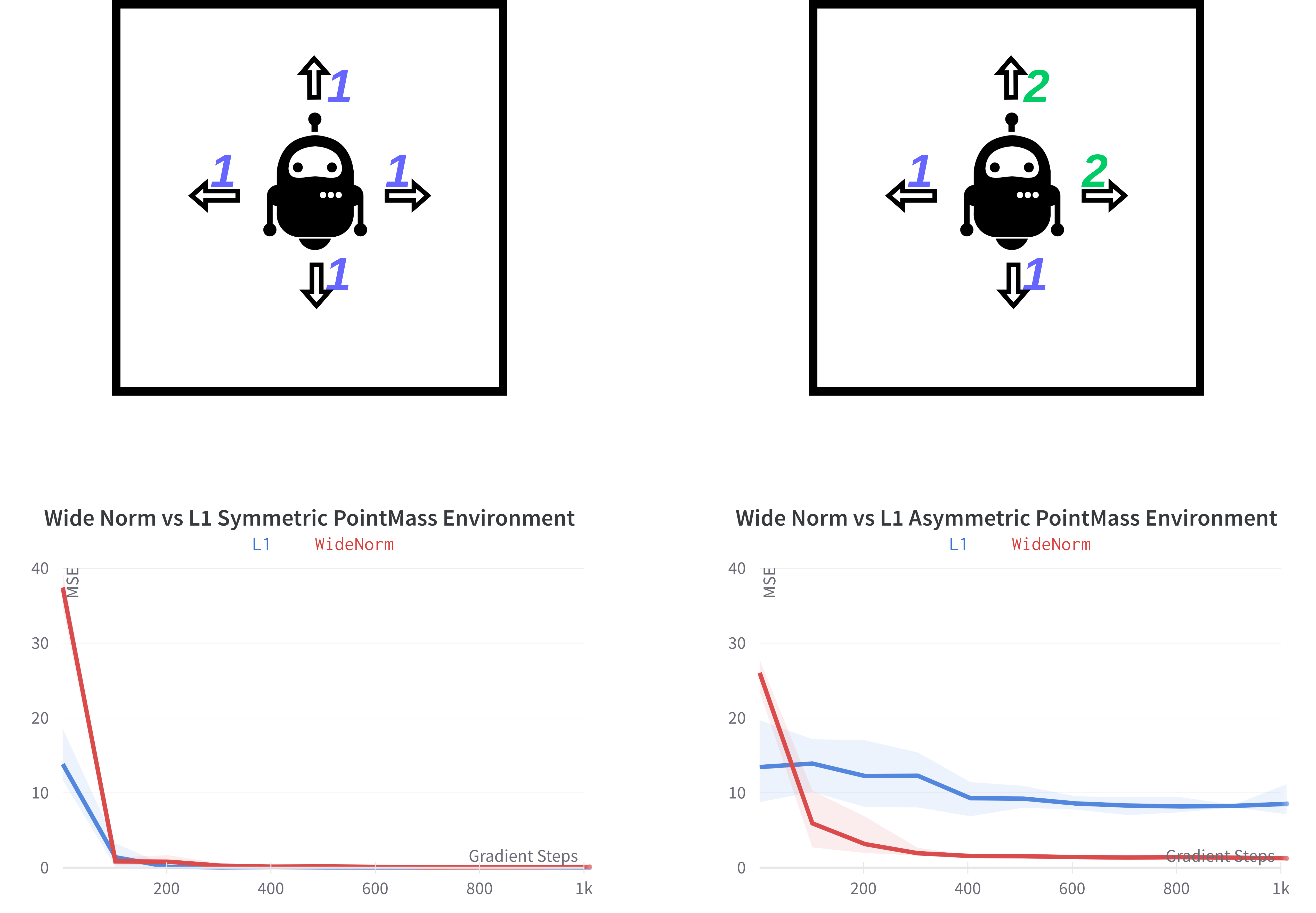}
    \caption{MSE respect to the true $d_{MAD}$.}
    \label{fig:MSE}
\end{figure}

In Figure \ref{fig:MSE} we can observe that $L1$ is unable to approximate the MAD in asymmetric environments ( Asymmetric PointMass Environment) converging to $\min(d_{MAD}(s_i, s_j), d_{MAD}(s_j, s_i))$ while $WideNorm$ approximate the asymmetric $d_{MAD}$ correctly.

\begin{figure}
    \centering
    \includegraphics[width=\textwidth]{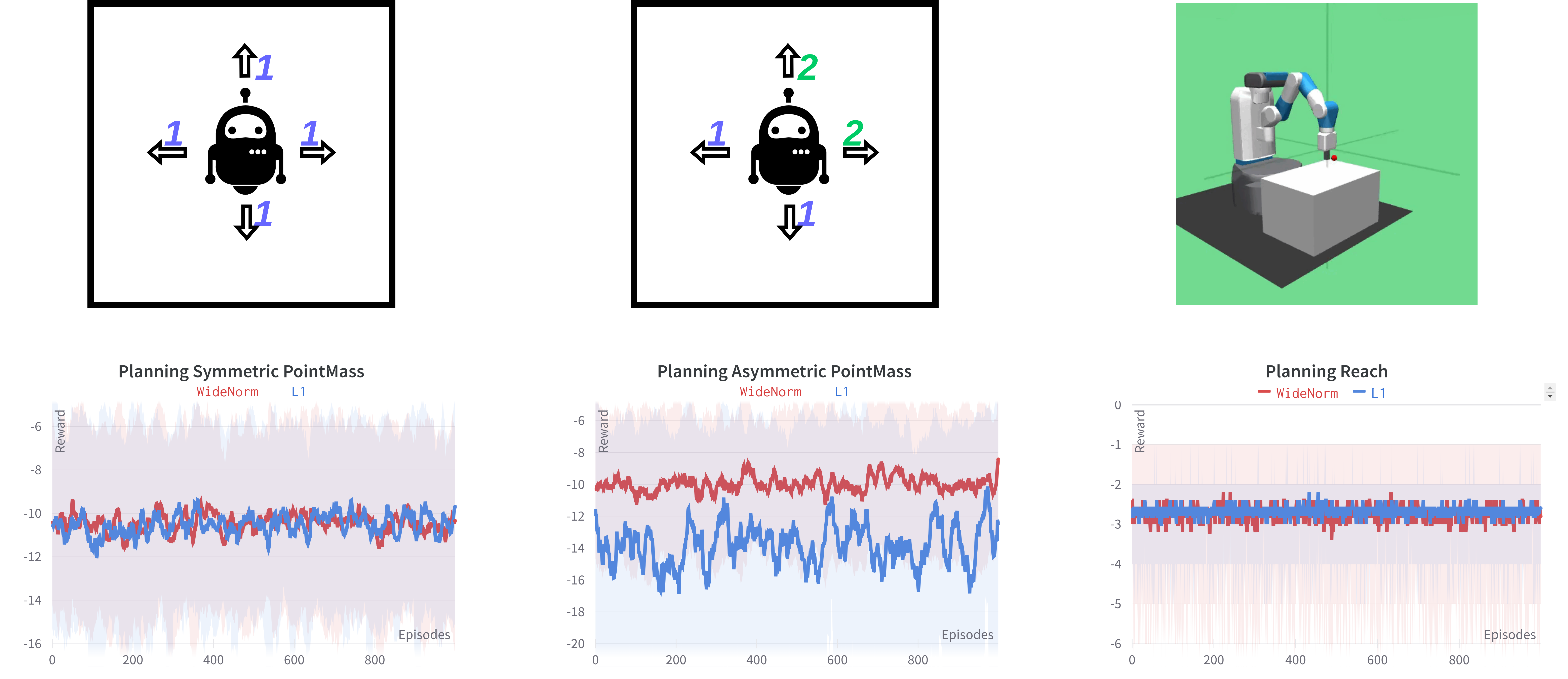}
    \caption{Comparison of WideNorms against P-norms on symmetric and asymmetric environments.}
    \label{fig:Planning}
\end{figure}

In Figure \ref{fig:Planning} we report the performance of the planning on the embedded space using the learned transition model. We can observe that in symmetric environments ( Planning Symmetric PointMass and Planning reach ) both $L1$ and $Widenorm$ have similar performance. But when we move to asymmetric environments ( Planning Asymmetric PointMass ) we can observe that $L1$ agent is unable to learn the true $d_{MAD}$ and as a consequence the planning performance degrades. In contrary $WideNorm$ is able to learn the true $d_{MAD}$ and outperforms $L1$ agent in planning.

The empirical evaluation demonstrates that symmetric norms such as the L1 norm $|| \cdot ||_1$ used in this experiment are unable to approximate the MAD in asymmetric environments converging to $\min(d_{MAD}(s_i, s_j), d_{MAD}(s_j, s_i))$ while WideNorms approximate the asymmetric $d_{MAD}$ correctly. 

We release the software to reproduce the results at the following repository: \href{https://github.com/lorenzosteccanella/SRL}{https://github.com/lorenzosteccanella/SRL} under the branch "\href{https://github.com/lorenzosteccanella/SRL/tree/NIPS-GCRL-Workshop}{NIPS-GCRL-Workshop}" and some notebooks that ease the understanding of the code under the branch "\href{https://github.com/lorenzosteccanella/SRL/tree/main}{main}".

\newpage

\bibliography{biblio}

\end{document}